\documentclass{article} % For LaTeX2e
\usepackage{times}

\usepackage{amsmath}
\usepackage{amssymb}
\usepackage{amsfonts}

\usepackage{graphicx} % more modern
\usepackage{subfigure}

\usepackage{natbib}

\newif\ifenoughspace
\enoughspacefalse

\newif\ifEnergybasedversion
\Energybasedversionfalse

\title{Implicit Density Estimation by Local Moment Matching to
Sample from Auto-Encoders}

\author{
Yoshua Bengio, Guillaume Alain, and Salah Rifai\\
Department of Computer Science and Operations Research\\
University of Montreal\\
Montreal, H3C 3J7 \\
}

\newcommand{\sigm}{\mathop{\mathrm{sigmoid}}}

\newcommand{\R}{\mathbb{R}}
\newcommand{\eqdef}{\stackrel{def}{=}}

\newcommand{\E}{\mathbb{E}}

\begin{document}

\maketitle
\vspace*{-2mm}
\begin{abstract}
  Recent work suggests that some auto-encoder variants do a good job of capturing the local manifold structure of the unknown data generating density.  This paper contributes to the mathematical understanding of this phenomenon and helps define better justified sampling algorithms for deep learning based on auto-encoder variants. We consider an MCMC where each step samples from a Gaussian whose mean and covariance matrix depend on the previous state, defines through its asymptotic distribution a target density. First, we show that good choices (in the sense of consistency) for these mean and covariance functions are the local expected value and local covariance under that target density. Then we show that an auto-encoder with a contractive penalty captures estimators of these local moments in its reconstruction function and its Jacobian. A contribution of this work is thus a novel alternative to maximum-likelihood density estimation, which we call local moment matching. It also justifies a recently proposed sampling algorithm for the Contractive Auto-Encoder and extends it to the Denoising Auto-Encoder.
\end{abstract}
\ifEnergybasedversion We also show how this training paradigm
  could in principle be applied to energy-based models in order to avoid
  computing the gradient of the partition function.  \fi

\vspace*{-2mm}
\section{Introduction}
\vspace*{-2mm}

Machine learning is about capturing aspects of the unknown distribution
from which the observed data are sampled (the {\em data-generating
distribution}). For many learning algorithms and in particular
in {\em manifold learning}, the focus is on identifying
the regions (sets of points) in the space of examples where
this distribution concentrates, i.e., which configurations of the
observed variables are plausible. 

Unsupervised {\em representation-learning} algorithms attempt to
characterize the data-generating distribution through the discovery of a
set of features or latent variables whose variations capture most of the
structure of the data-generating distribution.  In recent years, a number
of unsupervised feature learning algorithms have been proposed that are
based on minimizing some form of {\em reconstruction error}, such as
auto-encoder and sparse coding variants~\citep{Bengio-nips-2006-small,ranzato-07-small,
Jain-Seung-08-small,ranzato-08,VincentPLarochelleH2008-small,Koray-08-small,
Rifai+al-2011-small,Salah+al-2011-small,Dauphin-et-al-NIPS2011-small,gregor-nips-11-small}. 
An auto-encoder reconstructs the input through two stages, an encoder function $f$
(which outputs a learned representation $h=f(x)$ of an example $x$) and a decoder
function $g$, such that $g(f(x))\approx x$ for most $x$ sampled from the data-generating
distribution.
These feature learning algorithms
can be {\em stacked} to form deeper and more abstract representations. There
are arguments and much empirical evidence to suggest that when they are
well-trained, such {\em deep learning} algorithms~\citep{Hinton06,Bengio-2009,HonglakL2009,Salakhutdinov2009} can
perform better than their shallow counterparts, both in terms of learning features
for the purpose of classification tasks and for generating higher-quality 
samples.

Here we restrict ourselves to the case of continuous inputs
$x \in \R^d$ with the data-generating distribution being associated with an
unknown {\em target density} function, denoted $p$.
Manifold learning
algorithms assume that $p$ is concentrated in regions of lower 
dimension~\citep{Cayton-2005,Narayanan+Mitter-NIPS2010-short},
i.e., the training examples are by definition located very close to
these high-density manifolds.  In that context, the core objective of
manifold learning algorithms is to identify the local directions of variation,
such that small movement in input space along these directions stays on or near
the high-density manifold. 

Some important questions remain concerning many of these feature learning
algorithms. {\em What is their training criterion learning about the input
  density}? Do these algorithms implicitly learn about the whole density or
only some aspect? If they capture the essence of the target
density, then can we formalize that link and in particular exploit it
to {\em sample from the model}? This would turn these algorithms into
{\em implicit density} models, which only define a density indirectly, e.g., through
a generative procedure that converges to it.  These are the questions to which this
paper contributes.

A crucial starting point for this work is very recent work~\citep{Rifai-icml2012}
proposing a {\em sampling algorithm for Contractive Auto-Encoders}, detailed
in the next section. This algorithm was motivated on geometrical grounds,
based on the observation and intuition that the leading singular vectors
of the Jacobian of the encoder function specify those main directions
of variation (i.e., the tangent plane of the manifold,
the local directions that preserve the high-probability nature of training examples).
Here we make a formal link between the target density and models minimizing
reconstruction error through a contractive mapping, such as the
Contractive Auto-Encoder~\citep{Rifai+al-2011-small} and the Denoising
Auto-Encoder~\citep{VincentPLarochelleH2008-small}. This allows us to
justify sampling algorithms similar to
that proposed by~\citep{Rifai-icml2012}, and apply these ideas to
Denoising Auto-Encoders as well. 

We define a novel alternative to maximum
likelihood training, {\em local moment matching}, which
we find that Contractive and Denoising Auto-Encoders perform. This is achieved
by optimizing a criterion (such as a regularized reconstruction error) such that
the optimal learned reconstruction function (and its derivatives) provide estimators
of the local moments (and local derivatives) of the target density.
These local moments can be used to define an implicit density,
the asymptotic distribution of a particular Markov chain, which can also
be seen as corresponding to an {\em uncountable Gaussian mixture},
with one Gaussian component at each possible location in input space.
\ifEnergybasedversion 
As we show, the resulting sampling method could also
be applied to energy-based models, whose density is defined up to a
constant (the partition function), provided that the first and second
derivative of that density function (with respect to the input) can be
conveniently computed.  
\fi

The main novel contributions of this paper are the following.  First, we
show in Section~\ref{sec:CAE-DAE} 
that the Denoising Auto-Encoder with small Gaussian perturbations and
squared error loss is actually a Contractive Auto-Encoder whose contraction
penalty is the magnitude of the perturbation, making the theory developed
here applicable to both, and in particular extending the sampling procedure
used for the Contractive Auto-Encoder to the Denoising
Auto-Encoder as well. Second, we present in Section~\ref{sec:consistency} 
consistency arguments justifying the use
of the first and second estimated local moments to sample from a chain.
Such a sampling algorithm has successfully been used in~\citet{Rifai-icml2012} to sample from a Contractive
Auto-Encoder. With small enough steps, we show that the asymptotic
distribution of the chain has the same similar (smoothed) first and second
local moments as those estimated. Third, we show in Section~\ref{sec:min-rec} that non-parametrically minimizing
reconstruction error with a contractive regularizer yields a reconstruction
function whose value and Jacobian matrix estimate respectively the first
and second local moments, i.e., up to a scaling factor, are the right
functions to use in the Markov chain.  Finally, although the sampling
algorithm was already empirically verified in~\citet{Rifai-icml2012}, we
include in Section~\ref{sec:exp} an experimental validation for the case 
when the model is trained with the denoising criterion.

\vspace*{-2mm}
\section{Contractive and Denoising Auto-Encoders}
\label{sec:CAE-DAE}
\vspace*{-2mm}

The Contractive Auto-Encoder or CAE~\citep{Rifai+al-2011-small} 
is trained to minimize the following regularized
reconstruction error:
\begin{equation}
  {\cal L}_{CAE} = \E\left[ \ell(x,r(x))  + \alpha \left\|\frac{\partial f(x)}{\partial x}\right\|^2_F \right]
\label{eq:cae-crit}
\end{equation}
where $r(x)=g(f(x))$ and $||A||^2_F$ is the sum of the squares of the elements of $A$.
Both the squared loss $\ell(x,r)=\frac{1}{2}||x-r||^2$ and the
cross-entropy loss $\ell(x,r)=-x\log r -(1-x)\log(1-r)$ have been
used, but here we focus our analysis on the squared loss because
of the easier mathematical treatment it allows. Note that success of 
minimizing the above criterion
strongly depends on the parametrization of $f$ and $g$ and in particular
on the tied weights constraint used, with $f(x)=\sigm(W x + b)$
and $g(h)=\sigm(W^T h + c)$. The above regularizing term forces $f$ (as well
as $g$, because of the tied weights) to be contractive, i.e.,
to have singular values less than 1~\footnote{Note that an auto-encoder
without any regularization would tend to find many leading singular values near 1 in
order to minimize reconstruction error, i.e., preserve input norm
in all the directions of variation present in the data.}. Larger values of $\alpha$
yielding more contraction (smaller singular values) where it hurts reconstruction
error the least, i.e., in the local directions where there are only little or no
variations in the data.

The Denoising Auto-Encoder or DAE~\citep{VincentPLarochelleH2008-small}
is trained to minimize the following denoising criterion:
\begin{equation}
 {\cal L}_{DAE} = \E\left[ \ell(x,r(N(x))) \right]
\label{eq:dae-crit}
\end{equation}
where $N(x)$ is a stochastic corruption of $x$ and the expectation is over
the training distribution. Here we consider mostly
the squared loss and Gaussian noise corruption, again because it is easier to handle
them mathematically. In particular, if $N(x)=x+\epsilon$ with
$\epsilon$ a small zero-mean isotropic Gaussian noise vector of variance $\sigma^2$,
then a Taylor expansion around $x$ gives 
$r(x+\epsilon)\approx r(x)+\frac{\partial r(x)}{\partial x}\epsilon$, which
when plugged into ${\cal L}_{DAE}$ gives
\begin{eqnarray}
 {\cal L}_{DAE} &\approx& \E\left[ \frac{1}{2}\left(x-\left(r(x)+\frac{\partial r(x)}{\partial x}\epsilon\right)\right)^T
                                       \left(x-\left(r(x)+\frac{\partial r(x)}{\partial x}\epsilon\right)\right) \right] \nonumber \\
       &=& \frac{1}{2}\left(\E\left[\|x-r(x)\|^2\right] 
         - 2 E[\epsilon]^T \E\left[\frac{\partial r(x)}{\partial x}^T (x-r(x))\right] 
         + Tr\left(\E\left[\epsilon \epsilon^T\right]
              \E\left[\frac{\partial r(x)}{\partial x}^T\frac{\partial r(x)}{\partial x}\right]\right)\right) \nonumber \\
       &=& \frac{1}{2}\left(\E\left[\|x-r(x)\|^2\right] + \sigma^2 \E\left[\left\|\frac{\partial r(x)}{\partial x}\right\|^2_F\right]\right)
\label{eq:dae-crit-approx}
\end{eqnarray}
where in the second line we used the independance of the noise from $x$
and properties of the trace, while in the last line we used
 $\E\left[\epsilon \epsilon^T\right]=\sigma^2 I$
and $\E[\epsilon]=0$ by definition of $\epsilon$.
This derivation shows that {\em the DAE is also a Contractive Auto-Encoder} but
where the contraction is imposed explicitly on the whole reconstruction function
$r(\cdot)=g(f(\cdot))$ rather than on $f(\cdot)$ alone (and $g(\cdot)$ as a side
effect of the parametrization).

\vspace*{-2mm}
\subsection{A CAE Sampling Algorithm}
\label{sec:cae-sampling}
\vspace*{-2mm}

Consider the following coupled Markov chains with elements $M_t$ and $X_t$ respectively:
\begin{eqnarray}
 M_{t+1} &=& \mu(X_t) \nonumber \\
 X_{t+1} &=& M_{t+1} + Z_{t+1}.
\label{eq:2-chain}
\end{eqnarray}
where $M_t,X_t,\mu(X_t) \in \R^d$ and
$Z_{t+1}$ is a sample from a zero-mean Gaussian with covariance $\Sigma(X_t)$.

The basic algorithm for sampling from the CAE, proposed
in~\citet{Rifai-icml2012}, is based on the above Markov chain {\em
  operating in the space of hidden representation} $h=f(x)$, with
$\mu(h)=f(g(h))$ and $Z_{t+1}=\left(\frac{\partial f(x_t)}{\partial x_t}\right)
\left(\frac{\partial f(x_t)}{\partial x_t}\right)^T \varepsilon$, where $\varepsilon$ is
zero-mean isotropic Gaussian noise vector in $h$-space. This defines a chain of hidden
representations $h_t$, and the corresponding chain of input-space
samples is given by $x_t=g(h_t)$. Slightly better results are
obtained with this $h$-space than with the corresponding $x$-space
chain which defines 
$\mu(x)=g(f(x))$ and $Z_{t+1}=\left(\frac{\partial f(x_t)}{\partial x_t}\right)^T
\left(\frac{\partial f(x_t)}{\partial x_t}\right) \varepsilon$ where $\varepsilon$ is
zero-mean isotropic Gaussian noise in $x$-space. We conjecture
that this advantage stems from the fact that moves in $h$-space
are done in a more abstract, more non-linear space. 
\iffalse
Note also
that even though $g(f(\cdot))$ is trained as an auto-encoder
in $x$-space, it also means that $f(g(\cdot))$ is a good
auto-encoder in $h$-space, for the following reason. On most
training samples $x$ we have $g(f(x))\approx x$. Applying $f$
on both sides gives $f(g(f(x)))\approx f(x)$. Letting $h=f(x)$
means that $f(g(h))\approx h$ for high-probability $h$'s, i.e.,
with $h=f(x)$ and $x\sim p$. In what follows we mainly
study the $x$-space variant, but future work should formally
investigate the advantage of the $h$-space variant.
\fi

\vspace*{-2mm}
\section{Local Moment Matching as an Alternative to Maximum Likelihood}
\vspace*{-2mm}
\subsection{Previous Related Work}
\vspace*{-2mm}

Well-known manifold learning (``embedding'') algorithms include Kernel
PCA~\citep{Scholkopf98}, LLE~\citep{Roweis2000-lle},
Isomap~\citep{Tenenbaum2000-isomap}, Laplacian
Eigenmap~\citep{Belkin+Niyogi-2003}, Hessian
Eigenmaps~\citep{Donoho+Carrie-03}, Semidefinite
Embedding~\citep{Weinberger04a-small}, SNE~\citep{SNE-nips15-small} and
t-SNE~\citep{VanDerMaaten08-small} that were primarily developed and used
for data visualization through dimensionality reduction. These algorithms
optimize the hidden representation associated with training points 
in order to best preserve certain properties of an input-space
neighborhood graph.

The properties that we are interested in here are
the {\bf local mean} and {\bf local covariance}. They are defined as the mean and covariance of a density
restricted to a small neighborhood. For example, if we have lots of samples and we only
consider the samples around a point $x_0$, the local mean at $x_0$ would be estimated by the
mean of these neighbors and the local covariance at $x_0$ by the empirical covariance
among these neighbors.
There are previous machine learning algorithms that have been proposed to estimate
these local first and second moments by actually using local 
neighbors~\citep{Brand2003-small,Vincent-Bengio-2003-short,Bengio-Larochelle-NLMP-NIPS-2006-short}.
In Manifold Parzen Windows~\citep{Vincent-Bengio-2003-short} this is literally achieved by estimating
for each test point the empirical mean and empirical covariance of the neighboring
points, with a regularization of the empirical covariance that sets a floor value
for the eigenvalues of the covariance matrix. In Non-Local Manifold 
Parzen~\citep{Bengio-Larochelle-NLMP-NIPS-2006-short},
the mean and covariance are predicted by a neural network that takes $x_0$ as input
and outputs the estimated mean along with a basis for the leading eigenvectors
and eigenvalues of the estimated covariance. The predictor is trained to maximize
the likelihood of the near neighbors under the Gaussian with the predicted mean
and covariance parameters. Both algorithms are manifold learning
algorithms motivated by the objective to discover the local manifold structure
of the data, and in particular predict the manifold tangent planes around 
any given test point. Besides the computational difficulty of having to find the $k$ nearest neighbors
of each training example, these algorithms, especially Manifold Parzen Windows, heavily 
rely on the smoothness of the
target manifolds, so that there are enough samples to teach the model how and
where the manifolds bend.

Note that the term {\em local moment matching} was already used
by~\citet{Gerber-1982} in an actuarial context to match moments of a
discretized scalar distribution. Here we consider the more general problem
of modeling a multivariate density from data, by estimating the
first and second multivariate moments at every possible input point.

\vspace*{-2mm}
\subsection{A Sampling Procedure From Local Moment Estimators}
\vspace*{-2mm}

We first show that mild conditions suffice for the chain to 
converge~\footnote{This is also shown in~\citet{Rifai-icml2012}.}
and then that if local first and second moments have been estimated,
then one can define a plausible sampling algorithm based on a Markov chain
that exploits these local moments at each step. \\

{\bf Convergence of the Chain}\\

This Markov chain $X_1,X_2, \ldots X_t\ldots$ is 
precisely the one that samples a new point by sampling from the Gaussian
with these local first and second moments:
\begin{equation}
 X_{t+1}\sim N(\mu(X_{t}),\Sigma(X_{t}))
\label{eq:chain}
\end{equation}
as in Eq.\eqref{eq:2-chain}, but where we propose to choose
$\mu(X_t)-X_t$ proportional to 
the local mean at $x_0=X_t$ minus $X_t$, and $\Sigma(X_t)$
proportional to the local covariance at $x_0=X_t$. The functions $\mu(\cdot)$
and $\Sigma(\cdot)$ thus define a Markov chain. 

Let us sketch a proof that it converges under mild hypotheses, using the
decomposition into the chains of $M_t$'s and of $X_t$'s.  Assuming that
$\forall x,\;\; \mu(x)\in \cal B$ for some bounded ball $\cal B$, then \mbox{$M_t \in {\cal B}
\;\;\forall t$}. If we further assume that $\Sigma(X_t)$ is always full rank,
then there is a non-zero probability of jumping from any $M_t \in \cal B$
to any $M_{t+1} \in \cal B$, which is sufficient for ergodicity of the
chain and its convergence. Then if $M_t$'s converge, so do their noisy
counterparts $X_t$'s.\\

{\bf Uncountable Gaussian Mixture}\\

If the chain converges, let $\pi$ be the asymptotic distribution of the $X_t$'s.
It is interesting to note that $\pi$ satisfies the operator equation
\begin{equation}
  \pi(x) = \int \pi(\tilde{x}) {\cal N}(x;\mu(\tilde{x}),\Sigma(\tilde{x})) d\tilde{x}
\label{eq:mixture}
\end{equation}
where ${\cal N}(x;\mu(\tilde{x}),\Sigma(\tilde{x}))$ is the density of $x$ under
a normal multivariate distribution with mean $\mu(\tilde{x})$ and covariance $\Sigma(\tilde{x})$.
This can be seen as a kind of {\em uncountable Gaussian mixture} $\pi$ where 
the weight of each component $\tilde{x}$ is given by $\pi(\tilde{x})$ itself
and the functions $\mu(\cdot)$ and $\Sigma(\cdot)$ specify the mean and covariance
of each component.

\vspace*{-2mm}
\subsection{Consistency}
\label{sec:consistency}
\vspace*{-2mm}

From the point of view of learning,
what we would like is that the $\mu(\cdot)$ and $\Sigma(\cdot)$ functions
used in a sampling chain such as Eq.~\eqref{eq:chain} be such that they
yield an asymptotic density $\pi$ close to some target density $p$.
Because the Markov chain makes noisy finite steps, one would expect
that the best one can hope for is that $\pi$ be a smooth approximation
of $p$.

What we show below is that, in the asymptotic regime
of very small steps (i.e. $\Sigma(x)$ is small in magnitude),  a good choice is just
the intuitive one, where $\mu(x_0)-x_0 \propto \E[x|x_0]-x_0$ 
and $\Sigma(x_0) \propto Cov(x|x_0)$.

For this purpose, we formally define the local density $p_\delta(x|x_0)$
of points $x$ in the $\delta$-neighborhood of an $x_0$, as 
\begin{equation}
p_\delta(x|x_0)= \frac{p(x) 1_{||x-x_0||<\delta}}{Z(x_0)}
\label{eq:ptilde}
\end{equation}
where $Z(x_0)$ is the appropriate normalizing constant. Then
we respectively define the local mean and covariance around $x_0$ as simply being 
\begin{eqnarray}
 m_0 \eqdef \E[x|x_0] &=& \int x p_\delta(x|x_0) dx \nonumber \\
 C_0 \eqdef Cov(x|x_0) &=& \int (x-m_0)(x-m_0)^T p_\delta(x|x_0) dx.
\label{eq:mu0-C0}
\end{eqnarray}

Note that we have two scales here, the scale $\delta$ at which we take
the local mean and covariance, and the scale $\sigma=||\Sigma(x_0)||$
of the Markov chain steps. To prove consistency we assume here
that $\sigma \ll \delta$ and that both are small. Furthermore,
we assume that $\mu$ and $\Sigma$ are somehow calibrated, so that
the steps $\mu(x)-x$ are comparable in size to $\sigma$. This means
that $||\mu(x)-x|| \ll \delta$, which we use below.

We want to compute the local mean obtained when we follow the Markov chain,
i.e., in Eq.~\eqref{eq:ptilde} we choose a $p$ equal to $\pi$ 
(which is defined in Eq.~\eqref{eq:mixture}), and we obtain the following:
\begin{eqnarray}
 m_\pi \eqdef E_\pi[x|x_0] &=& \frac{1}{Z(x_0)}\int_x x \int_{\tilde{x}} p(\tilde{x}) {\cal N}(x;\mu(\tilde{x}),\Sigma(\tilde{x})) d\tilde{x}\; 1_{||x-x_0||<\delta} dx \nonumber \\
   &=& \frac{1}{Z(x_0)}\int_{\tilde{x}} p(\tilde{x}) \int_{||x-x_0||<\delta} x  {\cal N}(x;\mu(\tilde{x}),\Sigma(\tilde{x})) dx d\tilde{x}.
\label{eq:m-pi}
\end{eqnarray}
Because the Gaussian sample of the inner integral must be in a small region inside
the $\delta$-ball, the inner integral is approximately the Gaussian mean
if $\mu(\tilde{x})$ is in the $\delta$-ball, and 0 otherwise:
\begin{equation}
\int_{||x-x_0||<\delta} x  {\cal N}(x;\mu(\tilde{x}),\Sigma(\tilde{x})) dx
\approx
\mu(\tilde{x}) 1_{||\mu(\tilde{x})-x_0||<\delta}.\label{eq:first_moment_extracted}
\end{equation}
which gives
\begin{equation}
 m_\pi \approx \int_{\tilde{x}} \frac{p(\tilde{x})}{Z(x_0)} 1_{||\mu(\tilde{x})-x_0||<\delta} \mu(\tilde{x}) d\tilde{x}.
\label{eq:e-mu}
\end{equation}
Now we use the above assumptions, which give $||\mu(x)-x|| \ll \delta$, to conclude
that integrating under the region defined by $1_{||\mu(\tilde{x})-x_0||<\delta}$ is equivalent
to integrating under the region defined by $1_{||\tilde{x}-x_0||<\delta}$. Hence the
above approximation is rewritten
\begin{equation}
 m_\pi \approx \int_{\tilde{x}} \frac{p(\tilde{x})}{Z(x_0)} 1_{||\tilde{x}-x_0||<\delta} \mu(\tilde{x}) d\tilde{x}
\end{equation}
which by the definition of $\E[\cdot | x_0]$ gives the final result:
\begin{equation}
 m_\pi \approx \E[\mu(x)|x_0],
\end{equation}
It means that the local mean under the small-steps Markov chain is a local mean
of the chain's $\mu$'s. This justifies choosing $\mu(x)$ equal to the
local mean of the target density to be represented by the chain, so that
the Markov chain will yield an asymptotic distribution that has 
local moments that are close but smooth versions of those of the target density.

A similar result can be shown for the covariance by observing that the $x$ term in
Eq. (\ref{eq:first_moment_extracted}) produced the first moment of the Gaussian and
that the same reasoning would apply with $xx^T$ instead.

Alternatively, we can follow a shortened version of the above starting with
\begin{eqnarray}
 C_\pi & \eqdef & E_\pi[(x-m_\pi)(x-m_\pi)^T|x_0] \\
 &=& \frac{1}{Z(x_0)}\int_{\tilde{x}} p(\tilde{x}) \int_{||x-x_0||<\delta} (x-m_\pi)(x-m_\pi)^T  {\cal N}(x;\mu(\tilde{x}),\Sigma(\tilde{x})) dx
\end{eqnarray}
where
\begin{eqnarray}
\int_{||x-x_0||<\delta} (x-m_\pi)(x-m_\pi)^T  {\cal N}(x;\mu(\tilde{x}),\Sigma(\tilde{x})) dx \\
\approx
\left(\Sigma(\tilde{x})+(\mu(\tilde{x})-m_\pi)(\mu(\tilde{x})-m_\pi)^T \right) 1_{||\mu(\tilde{x})-x_0||<\delta}.\label{eq:second_moment_extracted}
\end{eqnarray}
By construction the magnitude of the covariance $\Sigma(\tilde{x})$ was made very small, so the following term vanishes
\begin{equation}
\int_{\tilde{x}} \frac{p(\tilde{x})}{Z(x_0)} 1_{||\tilde{x}-x_0||<\delta} \Sigma(\tilde{x}) d\tilde{x} \rightarrow 0
\end{equation}
and we are left with the desired result
\begin{equation}
C_\pi \approx \E\left[\left.(\mu(x)-m_\pi)(\mu(x)-m_\pi)^T\right|x_0\right].
\end{equation}

\vspace*{-2mm}
\subsection{Local Moment Matching By Minimizing Regularized Reconstruction Error}
\label{sec:min-rec}
\vspace*{-2mm}

We consider here an alternative to using nearest neighbors for estimating local moments,
by showing that {\em minimizing reconstruction error with a contractive penalty yields
estimators of the local mean and covariance.}

We start from a training criterion similar to the CAE's but penalizing the
contraction of the whole auto-encoder's reconstruction function, which is also equivalent
to the DAE's training criterion in the case of small Gaussian corruption noise 
(as shown in Eq.~\eqref{eq:dae-crit-approx}):
\begin{equation}
{\cal L}_{\rm global} = \int p(x_0) \left( \left\|x_0 - r(x_0)\right\|^2 + \alpha\left\|\frac{\partial r(x_0)}{\partial x_0}\right\|^2_F \right) dx_0
\label{eq:L-global}
\end{equation}
where $p$ is the target or training distribution. We prove that in a non-parametric 
setting (where $r$ is completely free), the optimal $r$ is such that the local 
mean $m_0$ is estimated by $r_0\eqdef r(x_0)$ while the local covariance 
$C_0$ is estimated\footnote{In practice, i.e., the parametric
case, there is no guarantee that the estimator be symmetric, but
this is easily fixed by symmetrizing it, i.e., using $\frac{J_0+J_0^T}{2}$.} 
by $J_0 \eqdef \left.\frac{\partial r}{\partial x}\right|_{x_0}$.

To find out what the auto-encoder estimates we follow an approach which has
already been used, e.g., to show that minimizing squared prediction error $\E[(f(X)-Y)^2]$ is
equivalent to estimating the conditional expectation, $f(X) \rightarrow \E[Y|X]$. For this
purpose we consider an asymptotic and non-parametric setting corresponding to the limit
where the number of examples goes to infinity (we actually minimize the
expected error) and the capacity goes to infinity: we allow the value $r_0$ and the 
derivative $J_0$ of $r$ at every point $x_0$, to be different, i.e., we
``parametrize'' $r(x)$ in every neighborhood around $x_0$ by
\begin{equation}
  r(x) = r(x_0) + \left.\frac{\partial r}{\partial x}\right|_{x_0} (x-x_0) = r_0 + J_0 (x-x_0)
\label{eq:r-local}
\end{equation}
which is like a Taylor expansion only valid in the neighborhood of $x$'s around $x_0$,
but where we actually consider $r_0$ and $J_0$ to be parameters free to be chosen
separately for each $x_0$.

Armed with this non-parametric formulation, we consider an infinity of such
neighborhoods and the local density $p_\delta(x|x_0)$ (Eq.~\ref{eq:ptilde})
which approaches a Dirac delta function in the limit of $\delta \rightarrow 0$, and
we rewrite ${\cal L}_{\rm global}$ as follows:
\begin{equation}
 {\cal L}_{\rm global} = \lim_{\delta \rightarrow 0} 
\int p(x_0) \left( \left(\int_x ||x - r(x)||^2 p_\delta(x|x_0) dx \right) + \alpha \left\|\frac{\partial r(x_0)}{\partial x_0}\right\|^2_F \right) dx_0.
\label{eq:L-global-dirac}
\end{equation}
The reason for choosing $p_\delta(x|x_0)$ that turns into a Dirac is that
the expectations of $x$ and $xx^T$ arising in the above inner integral
will give rise to the local mean $m_0$ and local covariance $C_0$.
If in the above equation we define $r(x)$ non-parametrically (as per Eq.~\eqref{eq:r-local}),
the minimum can be achieved by considering the separate minimization in each $x_0$ neighborhood
with respect to $r_0$ and $J_0$. We can express the local contribution to the loss at $x_0$ as
\begin{equation}
 {\cal L}_{\rm local}(x_0, \delta) = \int_x ||x - (r_0+J_0(x-x_0))||^2 p_\delta(x|x_0) dx + \alpha ||J_0||^2_F
\label{eq:L-local}
\end{equation}
so that
\begin{equation}
 {\cal L}_{\rm global} = \lim_{\delta \rightarrow 0} \int p(x_0) {\cal L}_{\rm local}(x_0, \delta) dx_0.
\end{equation}
We take the gradient of the local loss with respect to $r_0$ and $J_0$ and set it to 0 (detailed derivation in Appendix) to get
\begin{eqnarray}
 \frac{\partial {\cal L}_{\rm local}(x_0, \delta)}{\partial r_0} &=& 2(r_0 - m_0)+2J_0(m_0-x_0)\nonumber \\
 \frac{\partial {\cal L}_{\rm local}(x_0, \delta)}{\partial J_0} &=&  2\alpha J_0 -2\left(R-m_0 x_{0}^{T}-r_{0}(m_0-x_{0})^{T}\right) \nonumber \\
 & & \hspace{10pt} +2J_0\left(R-m_0 x_{0}^{T}-x_{0}m_0^{T}+x_{0}x_{0}{}^{T}\right).
\label{eq:gradients}
\end{eqnarray}
Solving these equations (detailed derivation in Appendix) gives us the solutions
\begin{eqnarray}
 r_0 &=& (I - J_0)m_0 + J_0 x_0 \nonumber \\
 J_0 &=& C_0 (\alpha I + C_0)^{-1}.
\label{eq:rJ-solutions}
\end{eqnarray}
Note that these quantities $m_0, C_0$ are defined through $p_\delta(x|x_0)$ so they depend implicitly on
$\delta$ and we should consider what happens when we take the limit $\delta \rightarrow 0$.

In particular, when $\delta \rightarrow 0$ we have that $\left\|C_0\right\| \rightarrow 0$ and we can see
from the solutions (\ref{eq:rJ-solutions}) that this forces $\left\|J_0\right\| \rightarrow 0$.
In a practical numerical application, we fix $\delta>0$ to be small and it becomes
interesting to see how these quantities relate asymptotically (in terms of $\delta$ decreasing).
In such a situation, we have the following asymptotically:\footnote{Here, to avoid confusion
with the overloaded $\sim$ notation for sampling, we instead use the $\asymp$ 
notation to denote that the ratio of any coefficient on the left with its corresponding coefficient on the right goes 
to 1 as $\delta \rightarrow 0$.}
\begin{eqnarray}
 r_0 & \asymp & m_0 \nonumber \\
 J_0 & \asymp & \alpha^{-1}C_0.
\label{eq:simple-rJ-solutions}
\end{eqnarray}

Thus, we have proved that {\em the value of the reconstruction and its Jacobian immediately
give us estimators of the local mean and local covariance, respectively.} 

\vspace*{-2mm}
\section{Experimental Validation}
\label{sec:exp}
\vspace*{-2mm}

In the above analysis we have considered the limit of a non-parametric case
with an infinite amount of data. In that limit, unsurprisingly,
reconstruction is perfect, i.e.,
$r(x)\rightarrow x$, $\E[x|x_0]\rightarrow x_0$,
and $||\frac{\partial r(x)}{\partial x}||_F\rightarrow 0$ and $||Cov(x|x_0)||_F\rightarrow 0$, 
at a speed that depends on the scale $\delta$, as we have seen
above. We do not care so much about the magnitudes but instead care about
the directions indicated by $r(x)-x$ (which indicate where to look for increases
in probability density) and by the singular vectors and
relative singular values of $\frac{\partial r(x)}{\partial x}$, which indicate what
directions preserve high density. In
a practical sampling algorithm such as described in Section~\ref{sec:cae-sampling},
one wants to take non-infinitesimal steps. Furthermore,
in the practical experiments
of ~\citet{Rifai-icml2012}, in order to get good generalization with a
limited training set, one typically works with a parametrized model which
cannot perfectly reconstruct the training set (but can generalize). 
This means that the learned reconstruction is not equal to the input,
even on training examples, and nor is the Jacobian of the reconstruction
function tiny (as it would be in the asymptotic non-parametric case).
The mathematical link between the two situations needs to be clarified
in future work, but a heuristic approach which we found to work well
is the following: control the scale of the Markov chain with a hyper-parameter
that sets the magnitude of the Gaussian noise (the variance of $\epsilon$
in Section~\ref{sec:cae-sampling}). 
That hyper-parameter can be optimized by visual inspection or by estimating
the log-likelihood of the samples, using a technique introduced
in~\citep{Breuleux+Bengio-2011} and also used in~\citet{Rifai-icml2012}.
The basic idea is to generate many samples from the model, train a
non-parametric density estimator (Parzen Windows) using these samples as a
training set, and evaluate the log-likelihood of the test set using that
density estimaor.  If the sample generator does not mix well, then some
test examples will be badly covered (far from any of the generated
samples), thus incurring a high price in log-likelihood. If the generator
mixes well but smoothes too much the true density, then the automatically
selected bandwidth of the Parzen Windows will be chosen larger, incurring
again a penalty in test log-likelihood.

\begin{figure*}
\begin{center}
\vspace*{-2mm}
    \includegraphics[width=0.95\textwidth]{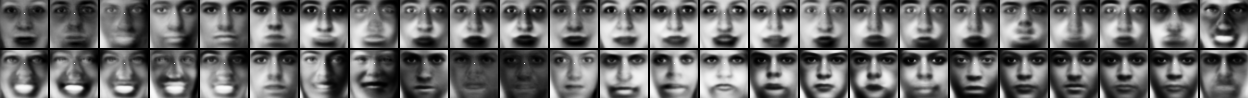}
    \includegraphics[width=0.95\textwidth]{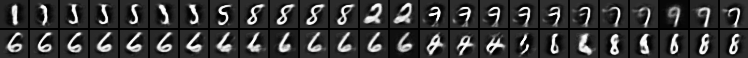}
\vspace*{-2mm}
\label{fig:samples}
\end{center}
\caption{Samples generated by a DAE trained on TFD (top 2 rows) and MNIST (bottom 2 rows).}
\end{figure*}

In Fig.~\ref{fig:samples}, we show samples of DAEs trained
and sampled similarly as in~\citet{Rifai-icml2012} on both MNIST digits images
and the Toronto Face Dataset (TFD)~\citep{Susskind2010}. These
results and those already obtained in~\citet{Rifai-icml2012}
confirm that the auto-encoder trained either as a CAE or a DAE
estimates local moments
that can be followed in a Markov chain to generate likely-looking
samples (and which have been shown quantitatively to be of high
quality in~\citet{Rifai-icml2012}).
%SAMPLING RESULTS WITH DAE: TO BE ADDED.

\ifEnergybasedversion
\section{Extension to Energy-Based Models}
\fi

\vspace*{-2mm}
\section{Conclusion}
\vspace*{-2mm}

This paper has contributed a novel approach to modeling densities: indirectly,
through the estimation of local moments. It has shown that local moments can
be estimated by auto-encoders with contractive regularization. It has justified
a sampling algorithm based on a simple Markov chain when estimators of the local 
moments are available. Whereas auto-encoders are unsupervised learning algorithms
that have been known for many decades, it has never been clear if they captured
everything that can be captured from a distribution. For the first time, this
paper presents a theoretical justification showing 
that they do implicitly perform density estimation
(provided some appropriate regularization is used, and assuming
the training criterion can be minimized).
This provides a more solid footing to the recently proposed algorithm
for sampling Contractive Auto-Encoders~\citep{Rifai-icml2012} and opens the door to other
related learning and sampling algorithms. In particular, it shows that this
sampling algorithm can be applied to Denoising Auto-Encoders as well.

An interesting advantage of modeling data through such training criteria is that 
there is no need to estimate an untractable partition function or its gradient,
and that there is no difficult inference problem associated with these types
of models either.
Future work following up on this paper
should try to answer the more difficult mathematical questions of 
what happens (e.g., with the consistency arguments presented here)
if the Markov chain steps are not tiny, and when we consider a
learner that has parametric constraints, rather than the asymptotic
non-parametric limit considered here.
We believe that the approach presented here can also be applied to
energy-based models (for which the free energy can be computed), and that
the local moments are directly related to the first and second derivative
of the estimated density. Future work should clarify that relationship,
possibly giving rise to new sampling algorithms for energy-based models
(since we have shown here that one can sample from the estimated
density if one can compute the local moments).
Finally, it would be interesting to extend this work in the direction of
analysis that explicitly takes into account the decomposition of the
auto-encoder into an encoder and a decoder. Indeed, we have found
experimentally that sampling in the representation space gives better
results than sampling in the original input space, but a more solid
mathematical treatment of such algorithms is still missing.

\vspace*{-2mm}

{\small
%\bibliography{strings,strings-shorter,ml,myrefs,aigaion-shorter}
\bibliography{strings,ml,aigaion-shorter}

\begin{thebibliography}{}

\bibitem[Belkin and Niyogi(2003)Belkin and Niyogi]{Belkin+Niyogi-2003}
Belkin, M. and Niyogi, P. (2003).
\newblock Laplacian eigenmaps for dimensionality reduction and data
  representation.
\newblock {\em Neural Computation\/}, {\bf 15}(6), 1373--1396.

\bibitem[Bengio(2009)Bengio]{Bengio-2009}
Bengio, Y. (2009).
\newblock Learning deep architectures for {AI}.
\newblock {\em Foundations and Trends in Machine Learning\/}, {\bf 2}(1),
  1--127.
\newblock Also published as a book. Now Publishers, 2009.

\bibitem[Bengio {\em et~al.}(2006)Bengio, Larochelle, and
  Vincent]{Bengio-Larochelle-NLMP-NIPS-2006-short}
Bengio, Y., Larochelle, H., and Vincent, P. (2006).
\newblock Non-local manifold {P}arzen windows.
\newblock In {\em NIPS'2005\/}, pages 115--122. MIT Press.

\bibitem[Bengio {\em et~al.}(2007)Bengio, Lamblin, Popovici, and
  Larochelle]{Bengio-nips-2006-small}
Bengio, Y., Lamblin, P., Popovici, D., and Larochelle, H. (2007).
\newblock Greedy layer-wise training of deep networks.
\newblock In {\em NIPS'2006\/}.

\bibitem[Brand(2003)Brand]{Brand2003-small}
Brand, M. (2003).
\newblock Charting a manifold.
\newblock In {\em NIPS'2002\/}, pages 961--968. {MIT} Press.

\bibitem[Breuleux {\em et~al.}(2011)Breuleux, Bengio, and
  Vincent]{Breuleux+Bengio-2011}
Breuleux, O., Bengio, Y., and Vincent, P. (2011).
\newblock Quickly generating representative samples from an rbm-derived
  process.
\newblock {\em Neural Computation\/}, {\bf 23}(8), 2053--2073.

\bibitem[Cayton(2005)Cayton]{Cayton-2005}
Cayton, L. (2005).
\newblock Algorithms for manifold learning.
\newblock Technical Report CS2008-0923, UCSD.

\bibitem[Donoho and Grimes(2003)Donoho and Grimes]{Donoho+Carrie-03}
Donoho, D.~L. and Grimes, C. (2003).
\newblock Hessian eigenmaps: new locally linear embedding techniques for
  high-dimensional data.
\newblock Technical Report 2003-08, Dept. Statistics, Stanford University.

\bibitem[Gerber(1982)Gerber]{Gerber-1982}
Gerber, H. (1982).
\newblock On the numerical evaluation of the distribution of aggregate claims
  and its stop-loss premiums.
\newblock {\em Insurance : Mathematics and Economics\/}, {\bf 1}, 13--18.

\bibitem[Gregor {\em et~al.}(2011)Gregor, Szlam, and
  LeCun]{gregor-nips-11-small}
Gregor, K., Szlam, A., and LeCun, Y. (2011).
\newblock Structured sparse coding via lateral inhibition.
\newblock In {\em NIPS'2011\/}.

\bibitem[Hinton and Roweis(2003)Hinton and Roweis]{SNE-nips15-small}
Hinton, G.~E. and Roweis, S. (2003).
\newblock Stochastic neighbor embedding.
\newblock In {\em NIPS'2002\/}.

\bibitem[Hinton {\em et~al.}(2006)Hinton, Osindero, and Teh]{Hinton06}
Hinton, G.~E., Osindero, S., and Teh, Y. (2006).
\newblock A fast learning algorithm for deep belief nets.
\newblock {\em Neural Computation\/}, {\bf 18}, 1527--1554.

\bibitem[Jain and Seung(2008)Jain and Seung]{Jain-Seung-08-small}
Jain, V. and Seung, S.~H. (2008).
\newblock Natural image denoising with convolutional networks.
\newblock In {\em NIPS'2008\/}.

\bibitem[Kavukcuoglu {\em et~al.}(2009)Kavukcuoglu, Ranzato, Fergus, and
  {LeCun}]{Koray-08-small}
Kavukcuoglu, K., Ranzato, M.-A., Fergus, R., and {LeCun}, Y. (2009).
\newblock Learning invariant features through topographic filter maps.
\newblock In {\em CVPR'2009\/}.

\bibitem[Lee {\em et~al.}(2009)Lee, Grosse, Ranganath, and Ng]{HonglakL2009}
Lee, H., Grosse, R., Ranganath, R., and Ng, A.~Y. (2009).
\newblock Convolutional deep belief networks for scalable unsupervised learning
  of hierarchical representations.
\newblock In L.~Bottou and M.~Littman, editors, {\em Proceedings of the
  Twenty-sixth International Conference on Machine Learning (ICML'09)\/}. ACM,
  Montreal (Qc), Canada.

\bibitem[Narayanan and Mitter(2010)Narayanan and
  Mitter]{Narayanan+Mitter-NIPS2010-short}
Narayanan, H. and Mitter, S. (2010).
\newblock Sample complexity of testing the manifold hypothesis.
\newblock In {\em NIPS'2010\/}.

\bibitem[Ranzato {\em et~al.}(2007)Ranzato, Poultney, Chopra, and
  {LeCun}]{ranzato-07-small}
Ranzato, M., Poultney, C., Chopra, S., and {LeCun}, Y. (2007).
\newblock Efficient learning of sparse representations with an energy-based
  model.
\newblock In {\em NIPS'06\/}.

\bibitem[Ranzato {\em et~al.}(2008)Ranzato, Boureau, and {LeCun}]{ranzato-08}
Ranzato, M., Boureau, Y.-L., and {LeCun}, Y. (2008).
\newblock Sparse feature learning for deep belief networks.
\newblock In J.~Platt, D.~Koller, Y.~Singer, and S.~Roweis, editors, {\em
  Advances in Neural Information Processing Systems 20 (NIPS'07)\/}, pages
  1185--1192, Cambridge, MA. MIT Press.

\bibitem[Rifai {\em et~al.}(2011a)Rifai, Vincent, Muller, Glorot, and
  Bengio]{Rifai+al-2011-small}
Rifai, S., Vincent, P., Muller, X., Glorot, X., and Bengio, Y. (2011a).
\newblock Contracting auto-encoders: Explicit invariance during feature
  extraction.
\newblock In {\em ICML'2011\/}.

\bibitem[Rifai {\em et~al.}(2011b)Rifai, Mesnil, Vincent, Muller, Bengio,
  Dauphin, and Glorot]{Salah+al-2011-small}
Rifai, S., Mesnil, G., Vincent, P., Muller, X., Bengio, Y., Dauphin, Y., and
  Glorot, X. (2011b).
\newblock Higher order contractive auto-encoder.
\newblock In {\em ECML PKDD\/}.

\bibitem[Rifai {\em et~al.}(2011c)Rifai, Dauphin, Vincent, Bengio, and
  Muller]{Dauphin-et-al-NIPS2011-small}
Rifai, S., Dauphin, Y., Vincent, P., Bengio, Y., and Muller, X. (2011c).
\newblock The manifold tangent classifier.
\newblock In {\em NIPS'2011\/}.

\bibitem[Rifai {\em et~al.}(2012)Rifai, Bengio, Dauphin, and
  Vincent]{Rifai-icml2012}
Rifai, S., Bengio, Y., Dauphin, Y., and Vincent, P. (2012).
\newblock A generative process for sampling contractive auto-encoders.
\newblock In {\em ICML'2012\/}.

\bibitem[Roweis and Saul(2000)Roweis and Saul]{Roweis2000-lle}
Roweis, S. and Saul, L.~K. (2000).
\newblock Nonlinear dimensionality reduction by locally linear embedding.
\newblock {\em Science\/}, {\bf 290}(5500), 2323--2326.

\bibitem[Salakhutdinov and Hinton(2009)Salakhutdinov and
  Hinton]{Salakhutdinov2009}
Salakhutdinov, R. and Hinton, G. (2009).
\newblock Deep {B}oltzmann machines.
\newblock In {\em Proceedings of the Twelfth International Conference on
  Artificial Intelligence and Statistics (AISTATS 2009)\/}, volume~8.

\bibitem[Sch{\"o}lkopf {\em et~al.}(1998)Sch{\"o}lkopf, Smola, and
  M{\"u}ller]{Scholkopf98}
Sch{\"o}lkopf, B., Smola, A., and M{\"u}ller, K.-R. (1998).
\newblock Nonlinear component analysis as a kernel eigenvalue problem.
\newblock {\em Neural Computation\/}, {\bf 10}, 1299--1319.

\bibitem[Susskind {\em et~al.}(2010)Susskind, Anderson, and
  Hinton]{Susskind2010}
Susskind, J., Anderson, A., and Hinton, G.~E. (2010).
\newblock The {T}oronto face dataset.
\newblock Technical Report UTML TR 2010-001, U. Toronto.

\bibitem[Tenenbaum {\em et~al.}(2000)Tenenbaum, {de Silva}, and
  Langford]{Tenenbaum2000-isomap}
Tenenbaum, J., {de Silva}, V., and Langford, J.~C. (2000).
\newblock A global geometric framework for nonlinear dimensionality reduction.
\newblock {\em Science\/}, {\bf 290}(5500), 2319--2323.

\bibitem[{van der Maaten} and Hinton(2008){van der Maaten} and
  Hinton]{VanDerMaaten08-small}
{van der Maaten}, L. and Hinton, G.~E. (2008).
\newblock Visualizing data using t-sne.
\newblock {\em J. Machine Learning Res.}, {\bf 9}.

\bibitem[Vincent and Bengio(2003)Vincent and Bengio]{Vincent-Bengio-2003-short}
Vincent, P. and Bengio, Y. (2003).
\newblock Manifold {P}arzen windows.
\newblock In {\em NIPS'2002\/}. MIT Press.

\bibitem[Vincent {\em et~al.}(2008)Vincent, Larochelle, Bengio, and
  Manzagol]{VincentPLarochelleH2008-small}
Vincent, P., Larochelle, H., Bengio, Y., and Manzagol, P.-A. (2008).
\newblock Extracting and composing robust features with denoising autoencoders.
\newblock In {\em ICML 2008\/}.

\bibitem[Weinberger and Saul(2004)Weinberger and Saul]{Weinberger04a-small}
Weinberger, K.~Q. and Saul, L.~K. (2004).
\newblock Unsupervised learning of image manifolds by semidefinite programming.
\newblock In {\em CVPR'2004\/}, pages 988--995.

\end{thebibliography}
\bibliographystyle{natbib}
}

\appendix

\section{Derivation of the local training criterion gradient}

We want to obtain the derivative of the local training criterion,
\begin{equation}
 L_{\rm local} = \lim_{\delta \rightarrow 0} 
 \int_x ||x - (r_0+J_0(x-x_0)||^2 \tilde{p}_\delta(x|x_0) dx + \alpha ||J_0||^2_F.
\label{eq:L-local}
\end{equation}
We use the definitions
\begin{eqnarray*}
 \mu_0 &=& \E[x|x_0]=\int_x x \tilde{p}_\delta(x|x_0) dx  \nonumber \\
 R_0 &=& \E[xx^T|x_0] \nonumber \\
 C_0 &=& Cov(x|x_0)=\E[(x-\mu_0)(x-\mu_0)^T|x_0]=R-\mu_0\mu_0^T.
\end{eqnarray*}
We first expand the expected square error:
\begin{eqnarray*}
 MSE=\E[||x-(r_0+J_0(x-x_0))||^2|x_0] &=& 
 \E[ (x - (r_0+J_0(x-x_0))(x - (r_0+J_0(x-x_0))^T ] \nonumber \\
 &=& \E[ (x - r_0)(x-r_0)^T-2(x-r_0)^TJ_0(x-x_0)] \nonumber \\
 &=& + \E[ (x-x_0)^TJ_0^TJ_0(x-x_0)]. \nonumber
\end{eqnarray*}
Differentiating this with respect to $r_0$ yields 
\[
 \frac{\partial MSE}{\partial r_0} = -2(\mu_0 - r_0)+2J_0(\mu-x_0)
\]
corresponding to Eq.(25) of the paper.

For differentiating with respect to $J_0$, we use the trace properties
\begin{eqnarray*}
  ||A||^2_F &=& Tr(A A^T) \nonumber \\
  Tr(ABC)&=&Tr(BCA) \nonumber \\
  %\frac{\partial Tr(A A^T)}{\partial A} &=& 2 A \nonumber \\
  \frac{\partial Tr(A^TXAZ)}{\partial A} &=& XAZ+X^TAZ^T \nonumber \\
  \frac{\partial Tr(X A)}{\partial A} &=& X^T \nonumber \\
  \frac{\partial Tr(X A^T)}{\partial A} &=& X.
\end{eqnarray*}
We obtain for the regularizer,
\[
 \frac{\partial \alpha ||J_0||^2_F}{\partial J_0}=2\alpha J_0
\]
and for the MSE:
\begin{eqnarray*}
 \frac{\partial MSE}{\partial J_0} &=& -2\E\left[\frac{\partial}{\partial J_0}tr(J_0(x-x_{0})(x-r_{0})^{T})\right]+\E\left[\frac{\partial}{\partial J_0}tr(J_0^{T}J_0(x-x_{0})(x-x_{0})^{T})\right]\\
 & = & -2\E\left[(x-r_{0})(x-x_{0})^{T}\right]+2J_0\E\left[(x-x_{0})(x-x_{0})^{T}\right]\\
 & = & -2\left(R-\mu_0 x_{0}^{T}-r_{0}(\mu_0-x_{0})^{T}\right)+2J_0\left(R-\mu_0 x_{0}^{T}-x_{0}\mu_0^{T}+x_{0}x_{0}{}^{T}\right)
\end{eqnarray*}

% & = & -2\left(R-\mu_0 x_{0}^{T}-\left(\mu_0-J_0(\mu_0-x_{0})\right)(\mu_0-x_{0})^{T}\right)+2J_0\left(R-\mu_0 x_{0}^{T}-x_{0}\mu_0^{T}+x_{0}x_{0}{}^{T}\right)\\
% & = & -2\left(R-\mu_0\mu_0^{T}\right)-2J_0(\mu_0-x_{0})(\mu_0-x_{0})^{T}+2J_0\left(R-\mu_0 x_{0}^{T}-x_{0}\mu_0^{T}+x_{0}x_{0}{}^{T}\right)\\
% & = & -2\left(R-\mu_0\mu_0^{T}\right)+2J_0\left(R-\mu_0\mu_0^{T}\right)\\
% & = & -2\left(I-J_0\right)C_0

%Putting all this together, we obtain
%\[
% \frac{\partial (MSE+\alpha ||J_0||^2_F)}{\partial J_0} = 2 \alpha J_0 - 2\left(I - J_0\right) C_0
%\] 
%as per Eq.(13) of the main text.

\section{Detailed derivation of the minimizers of the local training criterion}

Starting with $r_0$, we solve when the gradient is 0 to obtain
\begin{eqnarray*}
 \frac{\partial MSE}{\partial r_0} & = & (\mu-r_{0})-J_0(\mu-x_{0}) = 0\nonumber \\
\mu-r_{0} & = & J_0(\mu-x_{0})\nonumber \\
r_{0} & = & \mu-J_0(\mu-x_{0}) \nonumber \\
r_{0} & = & (I-J_0)\mu+J_0x_{0}.\nonumber
\end{eqnarray*}

Substituting that value for $r_0$ into the expression for the gradient with respect to $J_0$, we get
\begin{eqnarray*}
 \frac{\partial MSE}{\partial J_0}
 & = & -2\left(R-\mu_0 x_{0}^{T}-\left(\mu_0-J_0(\mu_0-x_{0})\right)(\mu_0-x_{0})^{T}\right)+2J_0\left(R-\mu_0 x_{0}^{T}-x_{0}\mu_0^{T}+x_{0}x_{0}{}^{T}\right)\\
 & = & -2\left(R-\mu_0\mu_0^{T}\right)-2J_0(\mu_0-x_{0})(\mu_0-x_{0})^{T}+2J_0\left(R-\mu_0 x_{0}^{T}-x_{0}\mu_0^{T}+x_{0}x_{0}{}^{T}\right)\\
 & = & -2\left(R-\mu_0\mu_0^{T}\right)+2J_0\left(R-\mu_0\mu_0^{T}\right)\\
 & = & -2\left(I-J_0\right)C_0.
\end{eqnarray*}
Adding the regularizer term and setting the gradient to 0, we get
\begin{eqnarray*}
 \frac{\partial (MSE+\alpha ||J_0||^2_F)}{\partial J_0} &=& -2(I-J_0)C_0+2\alpha J_0 = 0\\
C_0 & = & J_0 C_0 + \alpha J_0 \\
C_0 & = & J_0 (C_0 + \alpha I) \\
J_0 & = & C_0(\alpha I + C_0)^{-1}
\end{eqnarray*}

which altogether gives us Eq.(26) from the main text:
\begin{eqnarray*}
 r_0 &=& (I - J_0)\mu_0 + J_0 x_0 \\
 J_0 &=& (\alpha I + C_0)^{-1} C_0
\end{eqnarray*}

Note that we can also solve for $\mu_0$
\[
\mu = (I-J_0)^{-1}(r_{0}-J_0x_{0})
\]
and for $C_0$:
\begin{eqnarray*}
(I - J_0) C_0 &=& \alpha J_0 \\
C_0 &=& \alpha (I - J_0)^{-1} J_0
\end{eqnarray*}

However, we still have to take the limit as $\delta \rightarrow 0$.
Noting that the magnitude of $C_0$ goes to 0 as $\delta \rightarrow 0$,
it means that $J_0$ also goes to 0 in magnitude. Plugging in the above equations
gives the final results:
\begin{eqnarray*}
r_0 &=& \mu_0 \\
C_0 &=& J_0
\end{eqnarray*}

\end{document}